\documentclass[letterpaper,10pt,conference]{ieeeconf}
\IEEEoverridecommandlockouts
\usepackage{iros}
\usepackage{bm}

\acrodef{slam}[SLAM]{simultaneous localization and mapping}
\acrodef{tamp}[TAMP]{task and motion planning}
\acrodef{urdf}[URDF]{unified robot description format}
\acrodef{pt}[\emph{pt}]{parse tree}
\acrodef{iou}[IoU]{intersection over union}
\acrodef{map}[mAP]{mean average precision}
\acrodef{mre}[RE]{Range Error}
\acrodef{ra10}[RA]{Range Accuracy}
\acrodef{cpdr}[CDR]{Collision Detection Rate}
\acrodef{mde}[CDE]{Collision Distance Error}

\acrodef{ctar}[CSR]{Collision-free Success Rate}
\acrodef{aan}[ARS]{Average Rearrangement Steps}
\acrodef{acn}[ACN]{Average Collision Number}

\acrodef{moco}[MoCo-Net]{Motion Constraint Prediction Network}

\let\oldnl\nl
\newcommand{\nonl}{\renewcommand{\nl}{\let\nl\oldnl}}
\newcolumntype{x}{>{\columncolor{MistyRose}}c}
\newcolumntype{y}{>{\columncolor{LightCyan1}}c}

\title{\Large \bf Learning Contextual Motion Constraints for Object Rearrangement Exploration}

\author{Lexing Zhang$^{1,2}$\quad{}Zeyu Zhang$^{2\dagger}$\quad{}Gang Zeng$^{1}$\quad{}Hangxin Liu$^{1\dagger}$ 
\thanks{$^\dagger$ Corresponding authors. zhangzeyu@bigai.ai, hx.liu@pku.edu.cn}
\thanks{%
    $^{1}$ State Key Laboratory of General Artificial Intelligence, School of Intelligence Science and Technology, Peking University.
    $^{2}$ State Key Laboratory of General Artificial Intelligence, Beijing Institute for General Artificial Intelligence (BIGAI).
}
}

\begin{document}

\maketitle
\thispagestyle{empty}
\pagestyle{empty}

\begin{abstract}
The motion of an articulated part in cluttered surroundings is constrained not only by an object's intrinsic joint limits but also by extrinsic obstacles. However, existing approaches primarily consider object articulation properties alone. Here we present Motion Constraint Prediction Network (MoCo-Net), a two-stream neural network that jointly predicts the continuous motion ranges of articulated parts and obstacle heatmaps highlighting potential collision regions from partial-view point cloud observation. By capturing both internal and external motion constraints, MoCo-Net produces more accurate feasible motion range prediction and obstacles recognition, improves generalization for novel objects in new scenes, and enables robots to construct a scene configuration space for efficient exploration of rearrangement actions. Our experiments show that MoCo-Net's two-stream design enhances each other to deliver more accurate predictions, which subsequently form a compact configuration space for a customized planner to efficiently rearrange the scene.
\end{abstract}

\section{Introduction} \label{sec:introduction}

Understanding the structure of articulated objects is fundamental to their effective manipulation. Accordingly, prior research has devoted substantial effort to reconstructing part-level geometry~\cite{pekelny2008articulated,heppert2023carto,guo2025articulatedgs} and decomposing objects based on their articulation~\cite{yi2018deep,kawana2021unsupervised,wang2024active}. Further progress has been made in estimating articulated joint parameters~\cite{wang2019shape2motion,chu2023command,deng2024articulate} and predicting motion priors~\cite{mo2021where2act,eisner2022flowbot3d,wu2025afforddp}, thereby enabling more informed manipulation planning~\cite{jiao2025integration}.

Because articulation is visually ambiguous, many studies rely on \textbf{sequential observations} of an object's part motions to infer its structure and latent articulation properties~\cite{jain2021screwnet,qian2022understanding,liu2025building}. \textbf{Active interaction and probing} provide an additional means of disambiguating articulation~\cite{lv2022sagci,nie2022structure,ma2023sim2real}. In contrast, another line of work seeks to infer articulation properties from \textbf{static observations}, using only a single view~\cite{jiang2022opd,zhang2023part,yu2024gamma} or a small set of views~\cite{wei2022self,liu2023building,mandi2025realcode}. Although fully recovering articulation properties from static data is inherently ill-posed, these approaches are more versatile because they avoid the need for dense interactions or long observation sequences, which are not always available.

\begin{figure}[!t]
    \centering
    \includegraphics[width=\linewidth]{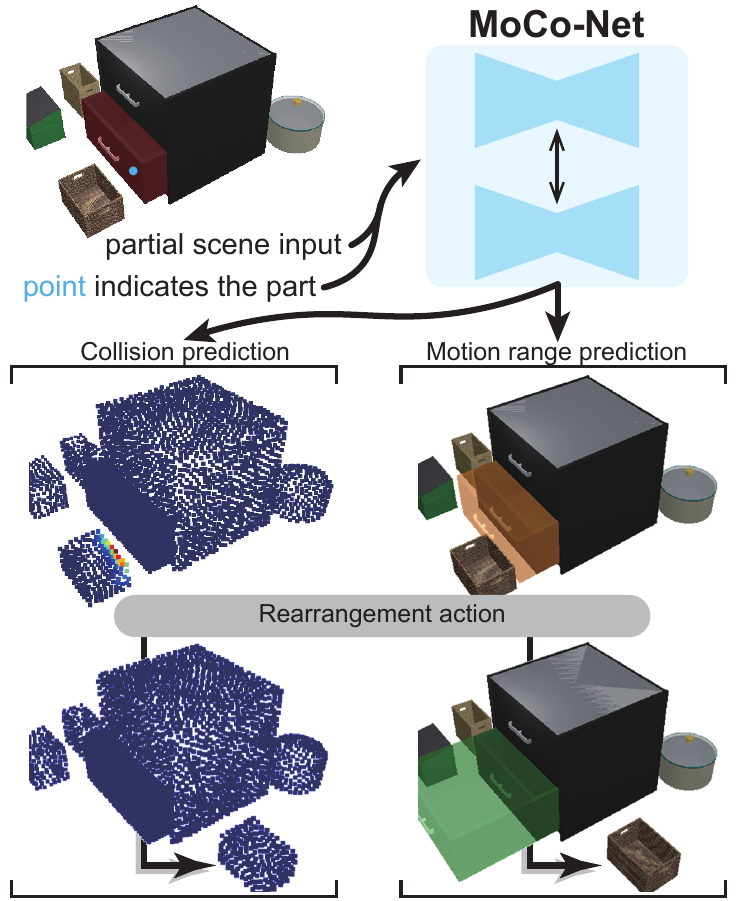}
    \caption{
        \textbf{An illustration of the proposed MoCo-Net.} Given a partial scene point cloud and a query point for the articulated part of interest, MoCo-Net predicts that part's feasible motion range and the heatmap highlighting surrounding regions where it may collide. Based on these predictions, an exploration algorithm could further compute rearrangement actions to effectively manipulate the part.
    }
    \label{fig:motivation}
\end{figure}

Despite these advances, most existing approaches overlook the influence of surrounding objects and contextual constraints on articulated parts. This is largely due to their object-centric modeling perspective; for instances, obtaining complete geometry from object reconstruction~\cite{mu2021sdf,heppert2023carto}, identifying articulated components by predicting movable parts~\cite{huang2014occlusion,geng2023gapartnet,sun2024opdmulti}, and inferring potential object interactions through affordance prediction~\cite{wu2022vat,li2023imagemanip,geng2023rlafford}. These approaches focus on understanding articulation of the object in isolation. Although some studies incorporate contextual cues such as support, containment, and non-penetration~\cite{han2021reconstructing,han2022scene} to improve object segmentation in scene reconstruction, they do not consider articulated objects or their parts in motion. As a result, they fail to capture how the feasibility of part motions or manipulation actions is shaped by constraints imposed by the surrounding environment.

To address this gap, we propose \ac{moco}, a two-stream neural network that jointly predicts the continuous feasible motion ranges of articulated parts in cluttered scenes and the obstacle heatmaps that restrict part movements. \ac{moco} is motivated by the insight that motion feasibility depends on both an object's intrinsic kinematic constraints and the extrinsic constraints imposed by surrounding obstacles. To capture both aspects, MoCo-Net performs scene-conditioned reasoning on a point-cloud input through two coordinated streams: one estimates collision-aware motion limits, while the other predicts an obstacle heatmap highlighting regions where the part is likely to collide, indicating constraints on part motion (\cref{fig:motivation}).

\begin{figure*}[!t]
    \centering
    \includegraphics[width=0.95\linewidth]{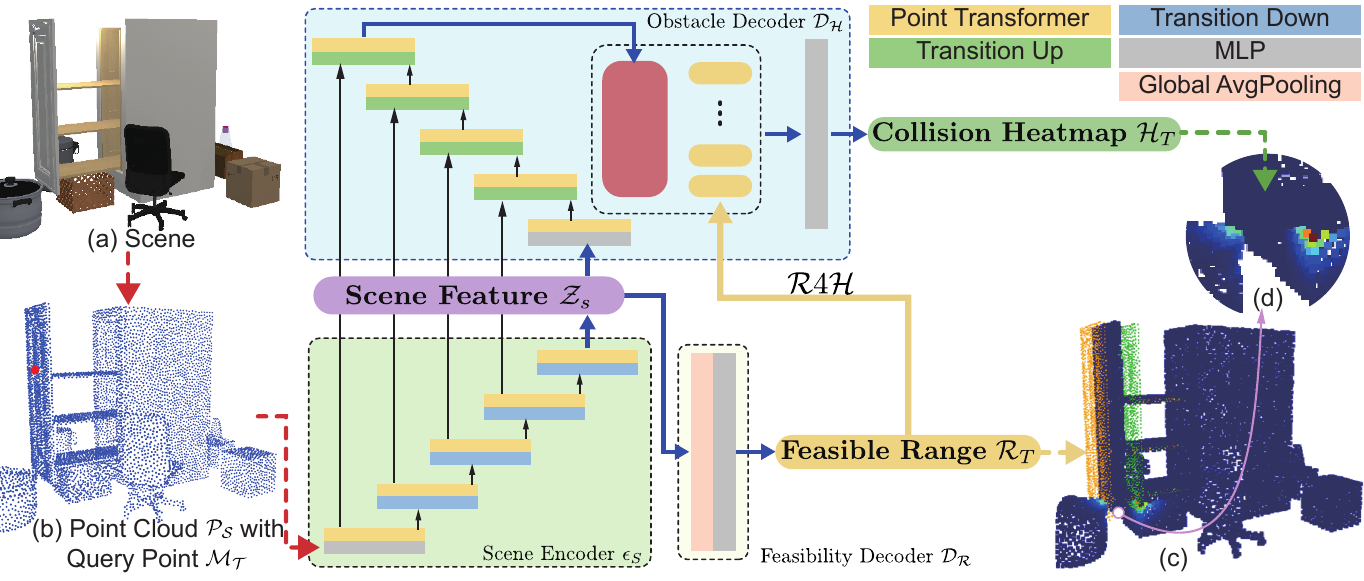}
    \caption{
        \textbf{Network architecture for feasible range and obstacle heatmap prediction.}
        (a) From the scene, we obtain (b) a partial scene point cloud $\mathcal{P}_{S}$ visualized in blue and a query point $\mathcal{M}_{T}$ visualized in red to specify the movable part.
        Our system predicts (c) the feasible range $\mathcal{R}_{T}$ and obstacle heatmap $\mathcal{H}_{T}$ for the movable part $T$.
        The feasible range is visualized through yellow and green points, representing upper and lower bounds respectively. 
        The obstacle heatmap is displayed using a color gradient, where warmer colors indicate higher collision likelihood. 
        In (d), we provide a zoomed view of highlighted areas in the heatmap to illustrate potential collision area.
    }
    \label{fig:network_architecture}
\end{figure*}

Our experiments show that these complementary predictions substantially improve accuracy and generalization, especially on unseen layouts and objects. Beyond prediction, the estimated feasible motion range provides a continuous parameterization of an articulated part's possible movements. This representation can be used to construct a scene configuration space that captures feasible prismatic and revolute part motions and supports subsequent analysis of rearrangement actions. A customized downstream planner is designed to leverage this configuration space to efficiently explore possible motions, assess action feasibility, and prioritize obstacle rearrangement based on predicted heatmaps, which reduces the need for exhaustive searches in high-dimensional action spaces. Because configuration space is a key concept in robot motion planning, MoCo-Net could offer a convenient and interpretable bridge between raw scene perception and actionable manipulation planning~\cite{jiao2021consolidating}.

The remainder of this paper is organized as follows: \cref{sec:method} introduces the proposed method. \cref{sec:planning} discuss the application of the feasible range and obstacle heatmap in planning. 
\cref{sec:experiments} describes the experimental setup and analysis of the results, and we conclude the paper in \cref{sec:conclusion}.

\section{MoCo-Net} \label{sec:method}

To better understand the environmental constraints on articulated object manipulation, we introduce a novel and challenging problem: inferring the feasible range of a movable part and its corresponding obstacle heatmap from a partial point cloud of a cluttered scene.

\subsection{Problem Formulation}
Consider a partial scene observation $S$ comprising $n$ points $\mathcal{P}_{S} = \{p^{S}_{1}, p^{S}_{2}, \cdots, p^{S}_{n}\}$, where each point $p^{S}_{i} \in \mathbb{R}^3$ represents a 3D coordinate. 
A movable part $T$ of interest is specified by a query point $\mathcal{M}_{T} \in \mathcal{P}_{S}$ from the scene point cloud.
As a sparse query signal, $\mathcal{M}_{T}$ conditions the prediction on the selected part when an object contains multiple movable parts.
Our objective is to predict two key elements for the movable part $T$.
The first is the feasible range $\mathcal{R}_{T} = [r_{\text{lower}}, r_{\text{upper}}]^{\top} \in \mathbb{R}^2$, indicating the maximum collision-free movement range relative to the current joint state.
For prismatic and revolute joints, the bounds are expressed as displacement and rotation increments, respectively.
The second is the obstacle heatmap $\mathcal{H}_{T} = \{h_1, h_2, \cdots, h_n\} \in [0, 1]^n$, subject to $\sum_{i=1}^{n} h_i = 1$, representing a collision likelihood distribution over scene points, highlighting where collisions are most likely to occur during the motion of part $T$.

We formulate the prediction of feasible range as $\mathcal{R}_{T} = f_{\theta}(\mathcal{P}_{S}, \mathcal{M}_{T})$, where $f_{\theta}$ is a predictor parameterized by $\theta$.
Similarly, for the more challenging obstacle heatmap prediction, we incorporate the predicted feasible range, expressing it as $\mathcal{H}_{T} = g_{\phi}(\mathcal{P}_{S}, \mathcal{M}_{T}, \mathcal{R}_{T})$, where $g_{\phi}$ is a predictor parameterized by $\phi$.
The problem is modeled as an optimization task, aiming to learn the optimal parameters $\theta$ and $\phi$ by minimizing the discrepancy between the predicted outputs ($\mathcal{R}_{T}$ and $\mathcal{H}_{T}$) and the ground truth ($\hat{\mathcal{R}}_{T}$ and $\hat{\mathcal{H}}_{T}$). The optimization objective is defined as:
\begin{equation}
    \small
    \begin{aligned}
        \min_{\theta, \phi} & \sum_{i=1}^{d} \
        \mathcal{L}_{\text{FR}}(\mathcal{R}_{T}^{(i)}, \hat{\mathcal{R}}_{T}^{(i)}) + \lambda \mathcal{L}_{\text{OH}}(\mathcal{H}_{T}^{(i)}, \hat{\mathcal{H}}_{T}^{(i)}) \\
        \mathcal{L}_{\text{FR}}(\mathcal{R}_{T}^{(i)}, \hat{\mathcal{R}}_{T}^{(i)})
        & = \text{MSE}(\mathcal{R}_{T}^{(i)}, \hat{\mathcal{R}}_{T}^{(i)})
        = \ ||f_{\theta}(\mathcal{P}_{S}^{(i)}, \mathcal{M}_{T}^{(i)}) - \hat{\mathcal{R}}_{T}^{(i)}||_2^2 \\
        \mathcal{L}_{\text{OH}}(\mathcal{H}_{T}^{(i)}, \hat{\mathcal{H}}_{T}^{(i)})
        & = \text{KL}(\mathcal{H}_{T}^{(i)}, \hat{\mathcal{H}}_{T}^{(i)}) \\
         = \sum_{j=1}^{n} \hat{\mathcal{H}}_{T}^{(i)}[j] & \log \frac{\hat{\mathcal{H}}_{T}^{(i)}[j]}{g_{\phi}(\mathcal{P}_{S}^{(i)}, \mathcal{M}_{T}^{(i)}, f_{\theta}(\mathcal{P}_{S}^{(i)}, \mathcal{M}_{T}^{(i)}))[j]},
    \end{aligned}
    \label{eq:optimization}
\end{equation}
where $d$ is the number of data samples, $\lambda$ is a balancing weight for the two loss terms, and $\text{MSE}$ and $\text{KL}$ denote the mean squared error and Kullback-Leibler divergence, respectively.
The MSE loss directly penalizes deviations in continuous motion range predictions, ensuring precise regression of motion limits. The KL divergence models the probability distribution over potential collision points, effectively capturing spatial constraints in the environment.
By coupling motion interval estimation with spatial collision distribution modeling, the formulation captures both intrinsic articulation limits and extrinsic environmental constraints within a unified predictive framework, enabling more accurate motion constraint estimation and facilitating downstream planning and manipulation tasks.

\subsection{Implementation and Model Training}

We implement the aforementioned predictors using a two-thread neural network, the architecture is illustrated in~\cref{fig:network_architecture}.
The first thread predicts the feasible range through the function $f_{\theta}$, where we optimize the parameters $\theta$. 
The second thread identifies movement-impeding obstacles through the function $g_{\phi}$, with $\phi$ as the learnable parameter.

The feasible range prediction thread takes in the partial scene point cloud $\mathcal{P}_{S}$ and the query point $\mathcal{M}_{T}$ and outputs the feasible range $\mathcal{R}_{T}$. 
This thread employs a Point Transformer-based~\cite{zhao2021point} scene encoder $\mathcal{E}_{S}$ to encode the point cloud input into a latent representation $\mathcal{Z}_{S}$, and a decoder $\mathcal{D}_{\mathcal{R}}$ that predicts the feasible range.
The obstacle heatmap prediction thread processes three inputs: the partial scene point cloud $\mathcal{P}_{S}$, the query point $\mathcal{M}_{T}$, and the feasible range $\mathcal{R}_{T}$.
This thread shares the scene encoder $\mathcal{E}_{S}$ with the prior thread and feeds local features from each encoder stage, along with the scene latent representation $\mathcal{Z}_{S}$ and feasible range $\mathcal{R}_{T}$, into decoder $\mathcal{D}_{\mathcal{H}}$ to predict the obstacle heatmap $\mathcal{H}_{T}$.

We jointly train both predictors by minimizing the loss function in \cref{eq:optimization}. 
The architecture $\mathcal{R}4\mathcal{H}$ creates a beneficial interdependence: gradients from the obstacle heatmap backpropagate to the feasible range predictor, enhancing its accuracy, while the feasible range serves as prior information that allows the obstacle heatmap predictor to focus on potential collision areas.

\section{Planning with Predicted Attributes} \label{sec:planning}

In this section, we describe how the predicted feasible range and obstacle heatmap facilitate articulated object manipulation in cluttered environments by efficiently exploring the scene configuration space.
The task is to manipulate a movable part $T_t$ by a desired amount $c_t$ in a cluttered environment, given a partial scene point cloud $\mathcal{P}_{S}$ and a point $\mathcal{M}_{T_t}$ specifying the part $T_t$.
The objective is to generate a rearrangement action sequence $\mathcal{A}$ to accomplish the task.

Our approach leverages predicted attributes to guide the exploration of the configuration space of a scene---an abstract space representing all possible object arrangements determined by their poses and geometries. 
Each configuration point corresponds to a unique arrangement, with free regions indicating collision-free configurations and forbidden regions representing collisions or constraint violations. 
The predicted feasible range reveals local structure while the obstacle heatmap identifies potential collisions, enabling safe and efficient exploration for planning.

\begin{algorithm}[t!]
    \small
    \caption{Planning for target articulated object}
    \label{alg:planning}
    \LinesNumbered
    \SetKwInOut{KIN}{Input}
    \SetKwInOut{KOUT}{Output}
    
    \SetKwFunction{Explore}{Explore}

    \KIN{Point $\mathcal{M}_{T_t}$, target state change $c_t$}
    \KOUT{A action sequence $\mathcal{A}$ to reach the goal: manipulate the part $T_t$ by $c_t$}

    \While {$c_t \neq 0$}{
        \While {True}{
            $\mathcal{P}_{S} \leftarrow \texttt{GetScenePointCloud}()$\;
            $\mathcal{R}_{T_t}, \mathcal{H}_{T_t} \leftarrow \texttt{MoCo-Net}(\mathcal{P}_{S}, \mathcal{M}_{T_t})$\;
            \If {$c_t$ in range $ \mathcal{R}_{T_t}$}{
                Break\;
            }
            $\mathcal{A}_{\text{explore}} \leftarrow \texttt{Explore}(\mathcal{M}_{T_t})$\; 
            $\mathcal{A} \leftarrow \mathcal{A} + \mathcal{A}_{\text{explore}}$\;
        }
        $a_{\text{manip}} \leftarrow \texttt{Rearrange}(T_t, c_t)$\;
        $\mathcal{A} \leftarrow \mathcal{A} + [a_{\text{manip}}]$\;
        $c_t \leftarrow c_t - a_{\text{manip}}.amount$\;
    }
\end{algorithm}
\begin{algorithm}[t!]
    \small
    \caption{Configuration Space Exploration}
    \label{alg:planing_explore}
    \LinesNumbered
    \SetKwProg{Fn}{Function}{:}{\KwRet}

    \Fn{\Explore{$\mathcal{M}_{T_k}$}}{
        $\mathcal{A} \leftarrow []$\;
        $\mathcal{P}_{S} \leftarrow \texttt{GetScenePointCloud}()$\;
        $\mathcal{R}_{T_k}^{(0)}, \mathcal{H}_{T_k}^{(0)} \leftarrow \texttt{MoCo-Net}(\mathcal{P}_{S}, \mathcal{M}_{T_k})$\;
        $\mathcal{M}_{T_u} \leftarrow \mathcal{P}_{S}[\argmax_{i} \mathcal{H}_{T_k}^{(0)}[i]]$\;
        $\mathcal{R}_{T_u}, \mathcal{H}_{T_u} \leftarrow \texttt{MoCo-Net}(\mathcal{P}_{S}, \mathcal{M}_{T_u})$\;
        \For {$c \in \mathcal{R}_{T_u}$}{
            $a_{\text{manip}} \leftarrow \texttt{Rearrange}(T_u, c)$\;
            $\mathcal{A} \leftarrow \mathcal{A} + [a_{\text{manip}}]$\;
            $\mathcal{P}_{S}^\text{new} \leftarrow \texttt{GetScenePointCloud}()$\;
            $\mathcal{R}_{T_k}, \mathcal{H}_{T_k} \leftarrow \texttt{MoCo-Net}(\mathcal{P}_{S}^\text{new}, \mathcal{M}_{T_k})$\;
            \If {range $\mathcal{R}_{T_k}$ contain range $\mathcal{R}_{T_k}^{(0)}$}{
                \KwRet{$\mathcal{A}$}\;
            }
            \Else{
                $a_{\text{manip}} \leftarrow \texttt{Rearrange}(T_u, -c)$\;
                $\mathcal{A} \leftarrow \mathcal{A} + [a_{\text{manip}}]$\;
            }
        }
        $\mathcal{A}_{\texttt{explore}} \leftarrow \texttt{Explore}(\mathcal{M}_{T_u})$\;
        \KwRet{$\mathcal{A} + \mathcal{A}_{\text{explore}}$}\;
    }
\end{algorithm}

As illustrated in~\cref{alg:planning}, the planning algorithm leverages the predicted feasible range and obstacle heatmap to facilitate articulated object manipulation. 
During each iteration, the algorithm predicts the feasible range $\mathcal{R}_{T_t}$ and obstacle heatmap $\mathcal{H}_{T_t}$ via function \texttt{MoCo-Net}.
When the target state lies within the feasible range, the algorithm executes a \texttt{Rearrange} action and updates the residual target state $c_t$. 
When obstacles prevent manipulation, the algorithm expands the feasible range of $T_t$ by exploring the configuration space via function \texttt{Explore} (see~\cref{alg:planing_explore}).

\begin{figure*}[!t]
    \centering
    \includegraphics[width=\linewidth]{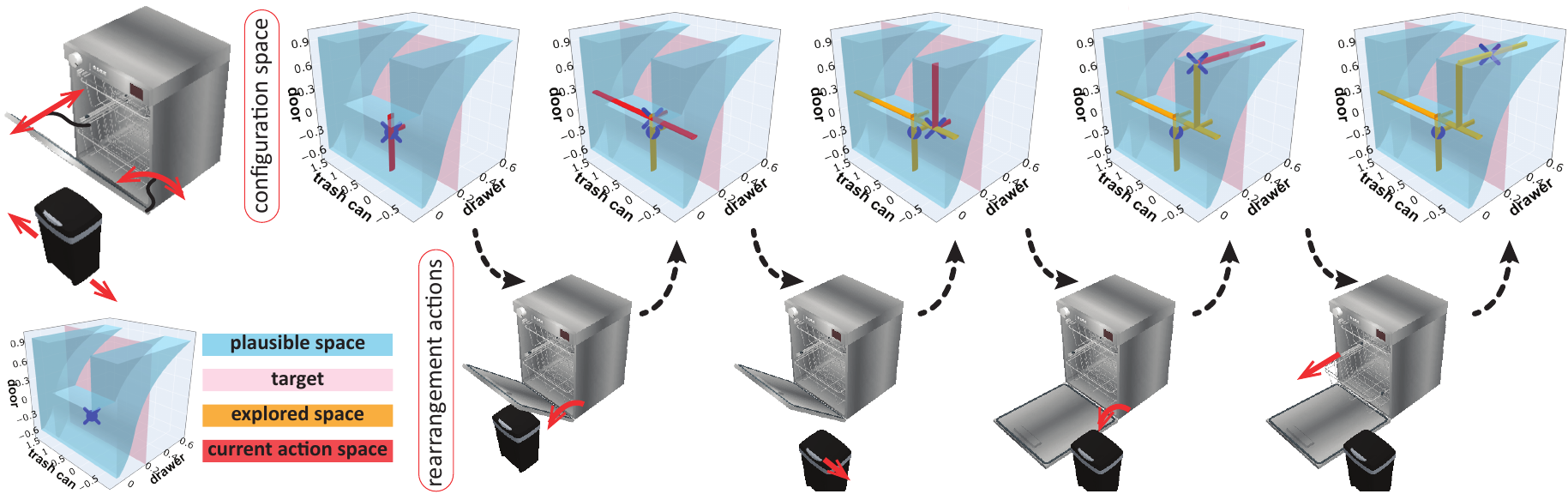}
    \caption{
        \textbf{Configuration space exploration for opening the dishwasher's bottom drawer by $0.2 m$.} The blue volume represents all feasible configurations of the scene (drawer joint position, door joint angle, trash can position simplified as 1D displacement). The blue circle marks the initial configuration and the blue cross marks the current configuration. Orange lines indicate previously explored configurations, while red lines show newly predicted feasible ranges where collision-free movement is possible. The algorithm efficiently expands within this space to reach the target, shown as the pink plane.
    }
    \label{fig:planning_example}
\end{figure*}

The exploration process begins by identifying the most likely collision point in $\mathcal{H}_{T_t}$.
We perform rearrangement actions on the movable part or obstacle $T_u$ corresponding to this point to expand the feasible range of $T_t$.
When necessary, the configuration space is recursively explored until the feasible range of $T_t$ is sufficiently expanded.

We demonstrate an example of exploring the configuration space using predicted attributes in a complex scene (see~\cref{fig:planning_example}). 
The scene consists of a dishwasher with three articulated parts (a door and two drawers) and a trash can. 
The door prevents the drawers from opening, while the trash can blocks the door. 
Our objective is to open the dishwasher's bottom drawer by $c_t = 0.2$ meter through the appropriate rearrangement of movable parts or obstacles.

Based on the initial observation, we predict the drawer's feasible range $\mathcal{R}_{\text{drawer}}^{(1)} = [0, 0.089]^{\top}$ and its obstacle heatmap $\mathcal{H}_{\text{drawer}}^{(1)}$. 
Since $c_t$ is not within range $\mathcal{R}_{\text{drawer}}^{(1)}$, the goal cannot be achieved directly. 
The heatmap $\mathcal{H}_{\text{drawer}}^{(1)}$ indicates that the door obstructs the drawer. 
As an articulated part, the door becomes our initial rearrangement target, with a predicted feasible range of $\mathcal{R}_{\text{door}}^{(1)} = [-0.510, 0.221]^{\top}$.
In the first action, we rearrange the door by opening it, thereby expanding the drawer's feasible range to $\mathcal{R}_{\text{drawer}}^{(2)} = [0, 0.097]^{\top}$. 
Despite this expansion, $c_t$ remains outside the range $\mathcal{R}_{\text{drawer}}^{(2)}$, necessitating further exploration of the configuration space. 
The updated heatmap $\mathcal{H}_{\text{drawer}}^{(2)}$ still identifies the door as an obstacle, but the door cannot be opened further in its current state.
Consequently, we recursively explore the configuration space, and the predicted obstacle heatmap $\mathcal{H}_{\text{door}}^{(2)}$ identifies the trash can as an obstacle blocking the door.
After the second rearrangement action of moving the trash can aside, the door's feasible range expands to $\mathcal{R}_{\text{door}}^{(3)} = [0, 0.711]^{\top}$. 
Subsequently, we perform the third rearrangement action of fully opening the door, further extending the drawer's feasible range to $\mathcal{R}_{\text{drawer}}^{(4)} = [0, 0.486]^{\top}$. 
Finally, as $c_t$ falls within the range $\mathcal{R}_{\text{drawer}}^{(4)}$, we open the drawer by $0.2$ meter to achieve the goal.

\section{Experiments} \label{sec:experiments}

To evaluate the effectiveness of our proposed method, we measure its performance across three aspects: feasible range prediction, obstacle heatmap prediction, and the impact of these predictions on downstream planning. 
Additionally, we conduct ablation studies to examine the architectural design of our model, including the joint training of the two predictors and leveraging predicted feasible ranges for obstacle heatmap prediction.

\subsection{Simulation and Data Generation}

\textbf{Simulation:} We adopt SAPIEN~\cite{xiang2020sapien} as our simulation environment and source objects from the ShapeNet dataset~\cite{chang2015shapenet} and Partnet-Mobility dataset~\cite{mo2019partnet}, which together provide a diverse collection of articulated objects across multiple categories. 
Our virtual scenes consist of target articulated objects placed with random joint states, surrounded by 0 to 15 randomly positioned obstacles. 
We focus on common interactive objects frequently encountered in daily environments, spanning a variety of household object categories such as tables, storage furniture, refrigerators, microwaves, \etc.

\textbf{Dataset:} $\mathcal{D} = \{(\mathcal{P}_{S}^{(i)}, \mathcal{M}_{T}^{(i)}, \hat{\mathcal{R}}_{T}^{(i)}, \hat{\mathcal{H}}_{T}^{(i)}, \hat{\mathcal{P}}_{C}^{(i)})\}$ comprises input point cloud $\mathcal{P}_{S}$, query point $\mathcal{M}_{T}$, ground truth feasible range $\hat{\mathcal{R}}_{T}$, obstacle heatmap $\hat{\mathcal{H}}_{T}$, and collision point set $\hat{\mathcal{P}}_{C}$.
To generate the input observation, we capture depth images from randomly positioned front-facing cameras. 
We reconstruct partial scene point clouds from these depth images, apply FPS sampling~\cite{eldar1997farthest}, and normalize the results to obtain $\mathcal{P}_{S}$. 
The query point $\mathcal{M}_{T}$ is extracted from the simulator by selecting a target movable part according to ground-truth part annotations and randomly sampling one point on that part.
The ground truth feasible range $\hat{\mathcal{R}}_{T}$ and obstacle heatmap $\hat{\mathcal{H}}_{T}$ are obtained by simulating the movement of the target movable part toward both bounds until collisions occur. 
The joint state changes at these collisions determine $\hat{\mathcal{R}}_{T}$. 
At the corresponding collision states, we record the impulse-filtered contact positions reported by SAPIEN between the moving part and obstacles as collision points $\hat{\mathcal{P}}_{C} = \{p^C_{1}, p^C_{2}, \cdots, p^C_{n_C}\}$.
Each collision is typically represented by one or two such points.
We then compute the distance from each point $p^{S}_{i}$ in the scene to its closest collision point $d_i = \min_{j} ||p^{S}_{i} - p^C_{j}||_2$. 
The obstacle heatmap $\hat{\mathcal{H}}_{T}$ is calculated by scaling the distances $d_i$ and applying the softmax function:
\begin{equation}
    \hat{h_i} = \text{softmax}(-\alpha d_i) = \frac{e^{-\alpha d_i}}{\sum_{j=1}^{n} e^{-\alpha d_j}}
    \label{eq:obstacle_heatmap_ground_truth}
\end{equation}
where $\alpha$ controls the heatmap's sensitivity to distance.

\begin{table*}[!ht]
    \centering
    \small
    \caption{
        \textbf{Quantitative comparison of feasible range and obstacle heatmap prediction.}
        $\downarrow$ indicates lower is better and $\uparrow$ indicates higher is better.
        \textit{Layout Var.} evaluates performance on scenes with seen objects but different layouts.
        \textit{Unseen Obj.} evaluates generalization to objects not present in training.
        The best results are highlighted in bold.
    }
    \resizebox{\linewidth}{!}{%
        \setlength{\tabcolsep}{3pt}
        \begin{tabular}{c cc cc cc cc}
            \toprule
            & \multicolumn{4}{c}{Feasible Range Prediction} 
            & \multicolumn{4}{c}{Obstacle Heatmap Prediction} \\
            \cmidrule(lr){2-5} \cmidrule(lr){6-9}
            Method 
            & \multicolumn{2}{c}{Range Accuracy $\uparrow$}
            & \multicolumn{2}{c}{Range Error $\downarrow$} 
            & \multicolumn{2}{c}{Collision Detection Rate $\uparrow$}
            & \multicolumn{2}{c}{Collision Distance Error $\downarrow$} \\
            \cmidrule(lr){2-3} \cmidrule(lr){4-5}
            \cmidrule(lr){6-7} \cmidrule(lr){8-9}
            & Layout Var. & Unseen Obj.
            & Layout Var. & Unseen Obj.
            & Layout Var. & Unseen Obj.
            & Layout Var. & Unseen Obj. \\
            \midrule
            GT-Sim                      & 0.433 & 0.445 & 0.305 & 0.300 & 0.511 & 0.490 & 0.210 & 0.196 \\
            Backbone-PointNet           & 0.460 & 0.402 & 0.158 & 0.181 & 0.685 & 0.495 & 0.106 & 0.129 \\
            Backbone-PointNet++         & 0.545 & 0.506 & 0.126 & 0.138 & 0.738 & 0.532 & 0.133 & 0.164 \\
            FR-Only                     & 0.604 & 0.566 & 0.114 & 0.128 & -     & -     & -     & -     \\
            OH-Only                     & -     & -     & -     & -     & 0.896 & 0.728 & 0.064 & 0.097 \\
            Joint                       & 0.645 & 0.590 & 0.106 & 0.119 & 0.903 & 0.739 & 0.064 & 0.095 \\
            Joint-H4R                   & 0.615 & 0.576 & 0.110 & 0.125 & 0.905 & 0.746 & 0.064 & 0.093 \\
            Joint-R4H (Ours)            & \bf{0.648} & \bf{0.600} & \bf{0.103} & \bf{0.118} & \bf{0.912} & \bf{0.757} & \bf{0.062} & \bf{0.091} \\
            \bottomrule
        \end{tabular}
    }
    \label{tab:fr_and_oh_experiment}
\end{table*}

\subsection{Feasible Range and Obstacle Heatmap Prediction}

\begin{figure*}[!t]
    \centering
    \includegraphics[width=0.97\linewidth]{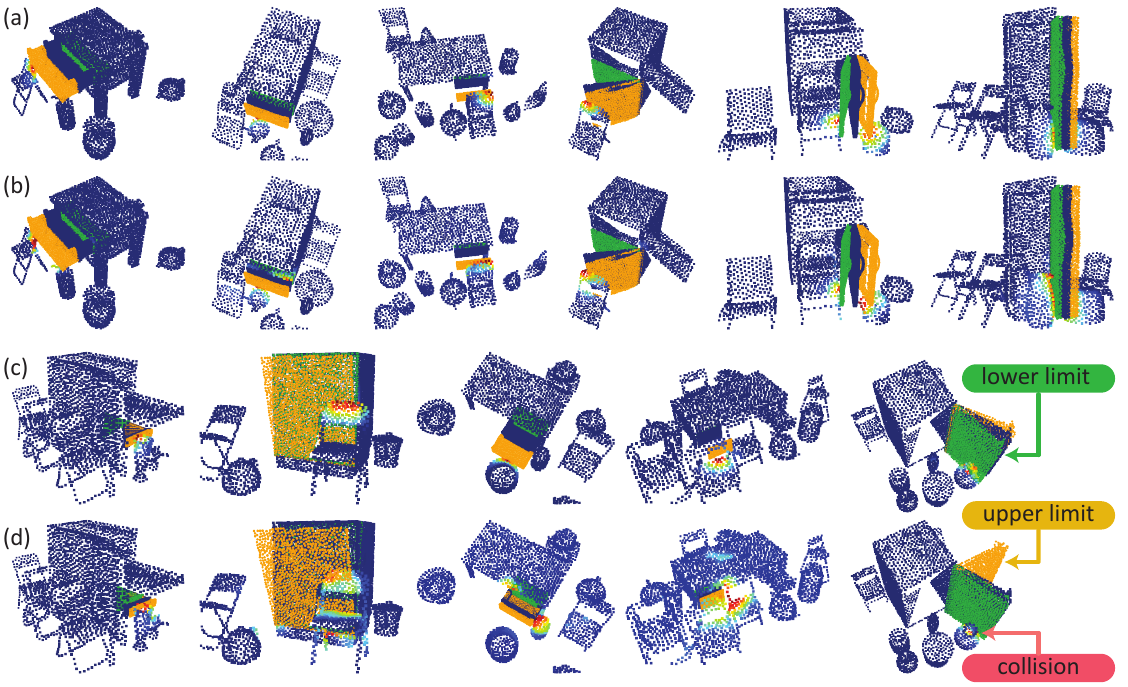}
    \caption{\textbf{Qualitative results of predicted joint motion ranges and potential collisions}. (a) and (c) show the ground-truth feasible ranges and collision heatmaps. (b) presents accurate predictions, while (d) illustrates less accurate and even false cases.}
    \label{fig:prediction_qual}
\end{figure*}

\textbf{Setup:} 
To systematically evaluate the performance of our method, we carefully curate the training and testing sets. The training set consists of 16,000 scenes, while the testing set is divided into two subsets: (i) \textit{Layout Var.}, comprising 5,000 scenes where target articulated objects and obstacles are sampled from the same object pool as the training set but with different object combinations, layouts, and viewpoints, and (ii) \textit{Unseen Obj.}, consisting of 5,000 scenes with entirely new target articulated objects, obstacles, layouts, and viewpoints. 
These subsets enable a comprehensive assessment of the generalization capability of the methods under varying scenarios.

Since no prior work directly addresses the problem we investigate, we compare our method with a set of carefully designed baselines and ablations, including a geometry-based simulation reference as well as variants that analyze backbone selection, the benefit of joint training, and cross-branch interaction in our framework.
\begin{itemize}
  \item \textit{GT-Sim}. 
  A baseline that computes feasible ranges and obstacle heatmaps by simulating part motion using ground-truth joint types, joint parameters, and part segmentation masks directly on the input point cloud.
  \item \textit{Backbone-PointNet} and \textit{Backbone-PointNet++}. 
  Variants of our full model where the backbone is replaced by PointNet~\cite{qi2017pointnet} and PointNet++~\cite{qi2017pointnet++}, respectively, while keeping all other components unchanged.
  \item \textit{FR-Only} and \textit{OH-Only}. 
  Single-task models that predict only the feasible range or only the obstacle heatmap, respectively, by removing the other prediction branch while retaining the same Point Transformer backbone.
  \item \textit{Joint}. 
  A shared-encoder model with two task-specific heads that jointly predicts the feasible range and obstacle heatmap, but without cross-branch conditioning.
  \item \textit{Joint-H4R}.
  A variant of \textit{Joint}, where the predicted obstacle heatmap is fed into the feasible range branch.
  \item \textit{Joint-R4H (Ours)}. 
  Our full model, where the predicted feasible range is used to guide obstacle heatmap estimation via $\mathcal{R}4\mathcal{H}$.
\end{itemize}
To ensure a fair comparison, our model and all ablations are trained under identical conditions, \ie, 100 epochs using the AdamW optimizer with a learning rate of $10^{-3}$ and a weight decay of $10^{-4}$.

\textbf{Metrics:}
We designed two metrics to evaluate the performance of the predicted feasible range: (i) \ac{mre} quantifies the normalized mean absolute error between the predicted and ground truth feasible range; (ii) \ac{ra10} measures the percentage of predictions within a 10\% error margin of the ground truth.
The \ac{mre} is defined as:
\begin{equation}
    \ac{mre} = \frac{1}{2|\mathcal{D}|} \sum_{i=1}^{|\mathcal{D}|} \frac{\Vert \mathcal{R}_{T}^{(i)} - \hat{\mathcal{R}}_{T}^{(i)} \Vert_1}{L_{\text{upper}}^{(i)} - L_{\text{lower}}^{(i)}}
    \label{eq:metric_mre}
\end{equation}
where $L_{\text{upper}}^{(i)}$ and $L_{\text{lower}}^{(i)}$ are the upper and lower joint limits of the target movable part in the $i$-th scene.
The \ac{ra10} metric is computed by:
\begin{equation}
    \begin{aligned}
        \ac{ra10} = \frac{1}{|\mathcal{D}|} \sum_{i=1}^{|\mathcal{D}|} 
        & \mathds{1}\left(\frac{\left|\mathcal{R}^{(i)}_T - \hat{\mathcal{R}}_{T}^{(i)}\right|}{L_{\text{upper}}^{(i)} - L_{\text{lower}}^{(i)}} \leq \begin{bmatrix}
 0.1\\0.1
\end{bmatrix}  \right) 
    \end{aligned}
    \label{eq:metric_ra10}
\end{equation}
where $\mathds{1}$ is an indicator function that returns 1 if the condition is satisfied and 0 otherwise.
The 10\% threshold represents a relative tolerance for normalized range prediction, providing a practical criterion to determine whether the predicted feasible range is sufficiently close to the ground truth.

We introduce two additional metrics to evaluate the performance of the predicted obstacle heatmap: the \ac{cpdr} and \ac{mde}. The \ac{cpdr} measures the percentage of successfully detected collision points, while the \ac{mde} quantifies the spatial accuracy by computing the distance between predicted and ground truth collision points.
To calculate \ac{cpdr}, we examine the averaged collision likelihood of neighboring points around each ground truth collision point. 
A collision point $p^{C}_{i,j}$ is considered detected if this averaged likelihood exceeds a threshold $\tau$. Formally, \ac{cpdr} is defined as:
\begin{equation}
    \ac{cpdr} = \frac{1}{|\mathcal{D}|} \sum_{i=1}^{|\mathcal{D}|} \frac{1}{|\hat{\mathcal{P}}_{C}^{(i)}|} \sum_{j=1}^{|\hat{\mathcal{P}}_{C}^{(i)}|}
\mathds{1}\left[g\left(p^{C}_{i,j},\mathcal{H}_T^{(i)}\right) \geq \tau \right]
    \label{eq:metric_cpdr}
\end{equation}
where $\hat{\mathcal{P}}_{C}^{(i)}$ represents the set of ground truth collision points in the $i$-th scene, $\mathcal{H}_T^{(i)}$ is the predicted obstacle heatmap, and $g\left(\cdot, \cdot \right)$ computes the average collision likelihood of neighboring points around the $j$-th collision point.
The threshold $\tau$ is used to distinguish confidently predicted collision regions in the heatmap.
For \ac{mde}, we first identify detected collision points $\mathcal{P}^{(i)}_C$ by thresholding the predicted obstacle heatmap $\mathcal{H}_T^{(i)}$. Then we compute the average minimum distance from each ground truth collision point to the closest detected collision point.
The metric is then computed as:
\begin{equation}
    \ac{mde} = \frac{1}{|\mathcal{D}|} \sum_{i=1}^{|\mathcal{D}|} \frac{1}{\left|\mathcal{P}_C^{(i)}\right|}
\sum_{p^{C}_{i,j} \in \mathcal{P}^{(i)}_C} d_{\min}\left(p^{C}_{i,j}, \hat{\mathcal{P}}^{(i)}_C\right)
    \label{eq:metric_mde}
\end{equation}
where $d_{\min}\left(\cdot, \cdot\right)$ computes the distance from a detected collision point $p^{C}_{i,j}$ to the closest ground truth collision point.

\textbf{Results:}
\cref{tab:fr_and_oh_experiment} reports the quantitative evaluation results using the proposed metrics for feasible range prediction and obstacle heatmap estimation. Our method (\textit{Joint-R4H (Ours)}) achieves the best performance across all metrics and evaluation settings and performs well under \textit{Unseen Obj.}, indicating accurate feasible range prediction, precise potential collision area identification, and strong generalization ability.

The \textit{GT-Sim} baseline shows poor performance as it performs collision checking directly on the observed partial point cloud, where incomplete observations and sparse point sampling lead to inaccurate collision estimation during simulated motion.
Additionally, replacing the Point Transformer (\textit{Joint-R4H (Ours)}) backbone with PointNet (\textit{Backbone-PointNet}) or PointNet++ (\textit{Backbone-PointNet++}) leads to consistent performance degradation in both feasible range and obstacle heatmap prediction, although PointNet++ improves over PointNet. This suggests that the attention mechanism in Point Transformer more effectively captures spatial and structural relationships among points and object parts, which is crucial for motion constraint reasoning.

For ablation, compared with single-task training, \textit{Joint} outperforms both \textit{FR-Only} and \textit{OH-Only} on their respective tasks, indicating that jointly learning feasible range and obstacle heatmap yields more expressive shared representations. This demonstrates the superiority of the coupled formulation in \cref{eq:optimization}.
Our models \textit{Joint-R4H} consistently surpasses both \textit{Joint} and \textit{Joint-H4R}, highlighting the effectiveness of the proposed $\mathcal{R}4\mathcal{H}$ design. In contrast, feeding heatmap predictions back to the range branch (\textit{Joint-H4R}) does not bring similar benefits and slightly harms range estimation. We attribute this to the higher complexity of heatmap prediction: inaccurate heatmap estimates may introduce misleading signals, whereas the predicted feasible range provides a more stable motion prior for guiding collision reasoning.

\cref{fig:prediction_qual} presents several qualitative examples of our predictions.
When comparing (a) with (b), our method accurately predicts the feasible range and obstacle heatmap, closely aligning with the ground truth.
However, (c) and (d) illustrate challenging cases where our method misjudges the feasible motion limits or fails to identify obstacles, resulting in less accurate predictions.

\begin{figure*}[!t]
    \centering
    \includegraphics[width=\linewidth]{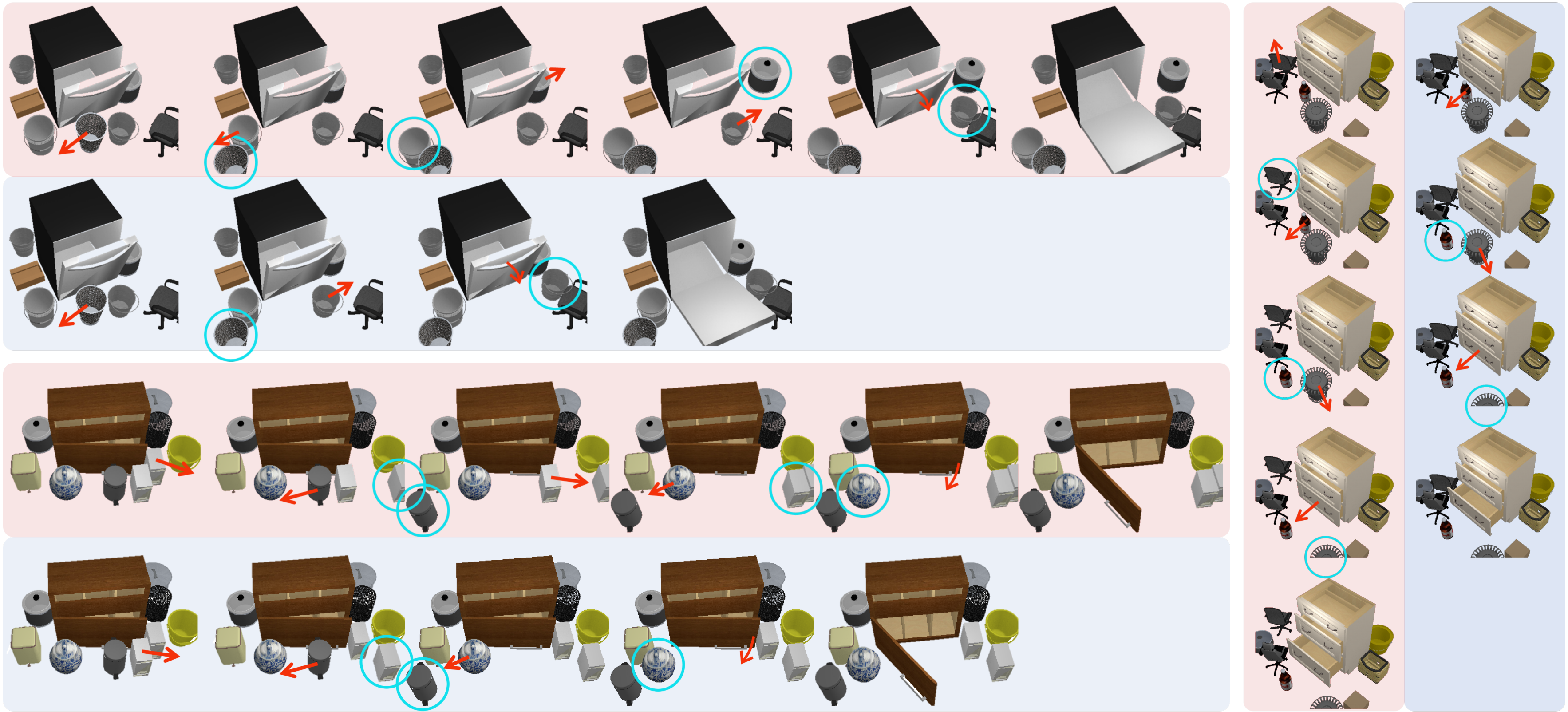}
    \caption{\textbf{Efficiency in computing rearrangement actions through the proposed configuration space exploration}. Rearrangement sequences shown in red are generated by the baseline method \textit{FR+Greedy}, which directly relocates the obstacle closest to the target part. In contrast, sequences shown in blue are produced by the proposed configuration space exploration algorithm, resulting in more efficient rearrangement actions.}
    \label{fig:rearrange}
\end{figure*}

\subsection{Configuration Space Exploration and Planning}

\textbf{Setup: } 
This experiment investigates how the predicted feasible range and obstacle heatmap facilitate planning for articulated object manipulation. 
We conduct experiments involving the manipulation of articulated objects by specified amounts. 
Each planning step involves a rearrangement action.
Specifically, we create 1,000 scenes, each containing a target articulated object surrounded by 0 to 15 obstacles.
The target manipulation amount is randomly sampled from the joint limit of the articulated part.
We compare planning with the predicted attributes (see~\cref{sec:planning}) against two baselines:
\begin{itemize}
    \item \textit{DirectManip}: Directly manipulates the target part, rearranging obstacles only when collisions occur.
    \item \textit{FR+Greedy}: Utilize the predicted feasible range to assess whether the goal is attainable and employ a greedy strategy to identify obstacles where the object nearest to the target part is considered as an obstacle.
\end{itemize}

\textbf{Metrics:}
We evaluate planning performance using three key metrics: (i) \ac{ctar} measures the percentage of instances where the articulated part successfully achieves the target state within 1\% of the intended manipulation amount without any collision; (ii) \ac{aan} quantifies planning efficiency by counting the average number of rearrangement actions required to reach the goal; Since each rearrangement step corresponds to a physical manipulation executed by the robot, this metric is also closely related to the execution time of the task;
 (iii) \ac{acn} evaluates the safety factor by measuring the average number of collisions occurring during planning execution. Together, these metrics provide a comprehensive assessment of planning effectiveness, efficiency, and safety.

\begin{table}[b!]
    \centering
    \small
    \caption{\textbf{Quantitative results of planning.} Arrows indicate the direction of improvement. The best results are highlighted in bold.}
    \resizebox{\linewidth}{!}{%
        \setlength{\tabcolsep}{11pt}
        \begin{tabular}{cccc}
            \toprule
            Method & \ac{ctar}$\uparrow$ & \ac{aan}$\downarrow$ & \ac{acn}$\downarrow$ \\ \hline \hline
            DirectManip & 0\% & 6.055 & 2.525 \\ 
            FR+Greedy & 80.0\% & 6.185 & 0.235 \\ 
            FR+OH (Ours) & \bf{82.4\%} & \bf{5.472} & \bf{0.201} \\ \hline
        \end{tabular}
    }%
    \label{tab:planning_experiment}
\end{table}

\textbf{Results:}
We present the quantitative results in~\cref{tab:planning_experiment}. 
The \textit{DirectManip} baseline exhibits a high \ac{acn} and a zero \ac{ctar}, indicating that directly manipulating the target part consistently leads to collisions and fails to achieve the goal safely.
In contrast, the \textit{FR+Greedy} baseline achieves a significantly higher \ac{ctar} of 80.0\% with a substantial reduction in \ac{acn} compared to \textit{DirectManip}, suggesting that our predicted feasible range enhances safety during manipulation.
Compared to the \textit{FR+Greedy} approach, our method reduces the arrangement steps by 0.713, highlighting the effectiveness of the predicted obstacle heatmap in identifying potential obstacles.
Overall, our method achieves a higher success rate while requiring fewer actions and producing fewer collisions during execution, demonstrating that our predicted motion constraints effectively guide configuration space exploration and lead to more efficient and safer planning in cluttered environments.

Qualitative examples of the rearrangement sequences are shown in \cref{fig:rearrange}.
Compared with the \textit{FR+Greedy} baseline, our method more accurately identifies obstacles and produces more efficient rearrangement sequences to achieve the goal.

\section{Conclusion} \label{sec:conclusion}
In this work, we introduced a novel approach for predicting feasible ranges and obstacle heatmaps from partial 3D point clouds to facilitate the manipulation of articulated objects in cluttered environments. 
Our method addresses a critical gap in existing work by modeling the continuous motion constraints on articulated objects imposed by both intrinsic articulated joints and extrinsic surrounding obstacles. 
Through extensive experiments, we demonstrated that our approach accurately predicts feasible ranges and generates meaningful obstacle heatmaps across diverse scenes and novel objects, outperforming the baselines. 
When integrated with planning algorithms, these predicted attributes significantly reduce the number of required actions and collisions and avoid collisions by efficiently guiding the exploration of high-dimensional configuration spaces.  
These results highlight the potential of our model to advance robotic manipulation capabilities in complex environments, offering a promising direction for future research in adaptive and context-aware robotics.

\balance
\setstretch{0.93}
\bibliographystyle{ieeetr}
\bibliography{reference}

@string {IJCV = "International Journal of Computer Vision (IJCV)"}

@string {TIP = "Transactions on Image Processing (TIP)"}

@string {CVPR = "Conference on Computer Vision and Pattern Recognition (CVPR)"}

@string {ICCV = "International Conference on Computer Vision (ICCV)"}

@string {ECCV = "European Conference on Computer Vision (ECCV)"}

@string {NIPS = "Advances in Neural Information Processing Systems (NeurIPS)"}

@string {ICLR = "International Conference on Learning Representations (ICLR)"}

@string {TRO = "IEEE Transactions on Robotics (T-RO)"}

@string {IROS = "IEEE/RSJ International Conference on Intelligent Robots and Systems (IROS)"}

@string {ICRA = "IEEE International Conference on Robotics and Automation (ICRA)"}

@string {RSS = "Robotics: Science and Systems (RSS)"}

@string {TOG = "ACM Transactions on Graphics (TOG)"}

@string {ThreeDV = "International Conference on 3D Vision (3DV)"}

@string {CGF = "Computer Graphics Forum"}

@inproceedings{mo2021where2act,
  title={Where2act: From pixels to actions for articulated 3d objects},
  author={Mo, Kaichun and Guibas, Leonidas J and Mukadam, Mustafa and Gupta, Abhinav and Tulsiani, Shubham},
  booktitle=ICCV,
  year={2021}
}

@inproceedings{eisner2022flowbot3d,
  title={FlowBot3D: Learning 3D Articulation Flow to Manipulate Articulated Objects},
  author={Eisner, Ben and Zhang, Harry and Held, David},
  journal=RSS,
  year={2022}
}

@inproceedings{han2021reconstructing,
  title={Reconstructing interactive 3d scenes by panoptic mapping and cad model alignments},
  author={Han, Muzhi and Zhang, Zeyu and Jiao, Ziyuan and Xie, Xu and Zhu, Yixin and Zhu, Song-Chun and Liu, Hangxin},
  booktitle=ICRA,
  year={2021}
}

@article{han2022scene,
  title={Scene reconstruction with functional objects for robot autonomy},
  author={Han, Muzhi and Zhang, Zeyu and Jiao, Ziyuan and Xie, Xu and Zhu, Yixin and Zhu, Song-Chun and Liu, Hangxin},
  journal=IJCV,
  volume={130},
  number={12},
  pages={2940--2961},
  year={2022}
}

@inproceedings{mu2021sdf,
  title={A-sdf: Learning disentangled signed distance functions for articulated shape representation},
  author={Mu, Jiteng and Qiu, Weichao and Kortylewski, Adam and Yuille, Alan and Vasconcelos, Nuno and Wang, Xiaolong},
  booktitle=ICCV,
  year={2021}
}

@inproceedings{wei2022self,
  title={Self-supervised neural articulated shape and appearance models},
  author={Wei, Fangyin and Chabra, Rohan and Ma, Lingni and Lassner, Christoph and Zollh{\"o}fer, Michael and Rusinkiewicz, Szymon and Sweeney, Chris and Newcombe, Richard and Slavcheva, Mira},
  booktitle=CVPR,
  year={2022}
}

@inproceedings{heppert2023carto,
  title={Carto: Category and joint agnostic reconstruction of articulated objects},
  author={Heppert, Nick and Irshad, Muhammad Zubair and Zakharov, Sergey and Liu, Katherine and Ambrus, Rares Andrei and Bohg, Jeannette and Valada, Abhinav and Kollar, Thomas},
  booktitle=CVPR,
  year={2023}
}

@inproceedings{chang2015shapenet,
  title={Shapenet: An information-rich 3d model repository},
  author={Chang, Angel X and Funkhouser, Thomas and Guibas, Leonidas and Hanrahan, Pat and Huang, Qixing and Li, Zimo and Savarese, Silvio and Savva, Manolis and Song, Shuran and Su, Hao and others},
  booktitle={arXiv preprint arXiv:1512.03012},
  year={2015}
}

@inproceedings{qi2017pointnet,
  title={Pointnet: Deep learning on point sets for 3d classification and segmentation},
  author={Qi, Charles R and Su, Hao and Mo, Kaichun and Guibas, Leonidas J},
  booktitle=CVPR,
  year={2017}
}

@inproceedings{qi2017pointnet++,
  title={Pointnet++: Deep hierarchical feature learning on point sets in a metric space},
  author={Qi, Charles Ruizhongtai and Yi, Li and Su, Hao and Guibas, Leonidas J},
  booktitle=NIPS,
  year={2017}
}

@article{yi2018deep,
  title={Deep part induction from articulated object pairs},
  author={Yi, Li and Huang, Haibin and Liu, Difan and Kalogerakis, Evangelos and Su, Hao and Guibas, Leonidas},
  journal=TOG,
  volume={37},
  number={6},
  pages={1--15},
  year={2018}
}

@inproceedings{mo2019partnet,
  title={Partnet: A large-scale benchmark for fine-grained and hierarchical part-level 3d object understanding},
  author={Mo, Kaichun and Zhu, Shilin and Chang, Angel X and Yi, Li and Tripathi, Subarna and Guibas, Leonidas J and Su, Hao},
  booktitle=CVPR,
  year={2019}
}

@inproceedings{wang2024active,
  title={Active Coarse-to-Fine Segmentation of Moveable Parts from Real Images},
  author={Wang, Ruiqi and Gadi Patil, Akshay and Yu, Fenggen and Zhang, Hao},
  booktitle=ECCV,
  year={2024}
}

@inproceedings{lv2022sagci,
  title={Sagci-system: Towards sample-efficient, generalizable, compositional, and incremental robot learning},
  author={Lv, Jun and Yu, Qiaojun and Shao, Lin and Liu, Wenhai and Xu, Wenqiang and Lu, Cewu},
  booktitle=ICRA,
  year={2022}
}

@inproceedings{zhang2023part,
  title={Part-level scene reconstruction affords robot interaction},
  author={Zhang, Zeyu and Zhang, Lexing and Wang, Zaijin and Jiao, Ziyuan and Han, Muzhi and Zhu, Yixin and Zhu, Song-Chun and Liu, Hangxin},
  booktitle=IROS,
  year={2023}
}

@inproceedings{wang2019shape2motion,
  title={Shape2motion: Joint analysis of motion parts and attributes from 3d shapes},
  author={Wang, Xiaogang and Zhou, Bin and Shi, Yahao and Chen, Xiaowu and Zhao, Qinping and Xu, Kai},
  booktitle=CVPR,
  year={2019}
}

@inproceedings{nie2022structure,
  title={Structure from action: Learning interactions for articulated object 3d structure discovery},
  author={Nie, Neil and Gadre, Samir Yitzhak and Ehsani, Kiana and Song, Shuran},
  booktitle={arXiv preprint arXiv:2207.08997},
  year={2022}
}

@inproceedings{chu2023command,
  title={Command-driven articulated object understanding and manipulation},
  author={Chu, Ruihang and Liu, Zhengzhe and Ye, Xiaoqing and Tan, Xiao and Qi, Xiaojuan and Fu, Chi-Wing and Jia, Jiaya},
  booktitle=CVPR,
  year={2023}
}

@inproceedings{geng2023rlafford,
  title={Rlafford: End-to-end affordance learning for robotic manipulation},
  author={Geng, Yiran and An, Boshi and Geng, Haoran and Chen, Yuanpei and Yang, Yaodong and Dong, Hao},
  booktitle=ICRA,
  year={2023}
}

@inproceedings{xiang2020sapien,
  title={Sapien: A simulated part-based interactive environment},
  author={Xiang, Fanbo and Qin, Yuzhe and Mo, Kaichun and Xia, Yikuan and Zhu, Hao and Liu, Fangchen and Liu, Minghua and Jiang, Hanxiao and Yuan, Yifu and Wang, He and others},
  booktitle=CVPR,
  year={2020}
}

@article{eldar1997farthest,
  title={The farthest point strategy for progressive image sampling},
  author={Eldar, Yuval and Lindenbaum, Michael and Porat, Moshe and Zeevi, Yehoshua Y},
  journal=TIP,
  volume={6},
  number={9},
  pages={1305--1315},
  year={1997}
}

@inproceedings{wu2022vat,
  title={VAT-Mart: Learning Visual Action Trajectory Proposals for Manipulating 3D ARTiculated Objects},
  author={Wu, Ruihai and Zhao, Yan and Mo, Kaichun and Guo, Zizheng and Wang, Yian and Wu, Tianhao and Fan, Qingnan and Chen, Xuelin and Guibas, Leonidas and Dong, Hao},
  booktitle=ICLR,
  year={2022}
}

@inproceedings{zhao2021point,
  title={Point transformer},
  author={Zhao, Hengshuang and Jiang, Li and Jia, Jiaya and Torr, Philip HS and Koltun, Vladlen},
  booktitle=ICCV,
  year={2021}
}

@inproceedings{jain2021screwnet,
    author = {Jain, Ajinkya and Lioutikov, Rudolf and Chuck, Caleb and Niekum, Scott},
    title = {ScrewNet: Category-Independent Articulation Model Estimation From Depth Images Using Screw Theory},
    booktitle = ICRA,
    year = {2021}
}

@inproceedings{jiao2021consolidating,
  title={Consolidating kinematic models to promote coordinated mobile manipulations},
  author={Jiao, Ziyuan and Zhang, Zeyu and Jiang, Xin and Han, David and Zhu, Song-Chun and Zhu, Yixin and Liu, Hangxin},
  booktitle=IROS,
  year={2021},
  organization={IEEE}
}

@article{jiao2025integration,
  title={Integration of Robot and Scene Kinematics for Sequential Mobile Manipulation Planning},
  author={Jiao, Ziyuan and Niu, Yida and Zhang, Zeyu and Wu, Yangyang and Su, Yao and Zhu, Yixin and Liu, Hangxin and Zhu, Song-Chun},
  journal=TRO,
  year={2025}
}

@inproceedings{liu2025building,
  title={Building interactable replicas of complex articulated objects via gaussian splatting},
  author={Liu, Yu and Jia, Baoxiong and Lu, Ruijie and Ni, Junfeng and Zhu, Song-Chun and Huang, Siyuan},
  booktitle=ICLR,
  year={2025}
}

@inproceedings{yu2024gamma,
  title={Gamma: Generalizable articulation modeling and manipulation for articulated objects},
  author={Yu, Qiaojun and Wang, Junbo and Liu, Wenhai and Hao, Ce and Liu, Liu and Shao, Lin and Wang, Weiming and Lu, Cewu},
  booktitle=ICRA,
  year={2024},
}

@inproceedings{jiang2022opd,
  title={Opd: Single-view 3d openable part detection},
  author={Jiang, Hanxiao and Mao, Yongsen and Savva, Manolis and Chang, Angel X},
  booktitle=ECCV,
  year={2022},
}

@inproceedings{mandi2025realcode,
  title={Real2Code: Reconstruct Articulated Objects via Code Generation},
  author={Mandi, Zhao and Weng, Yijia and Bauer, Dominik and Song, Shuran},
  booktitle=ICLR,
  year={2025},
}

@inproceedings{qian2022understanding,
  title={Understanding 3d object articulation in internet videos},
  author={Qian, Shengyi and Jin, Linyi and Rockwell, Chris and Chen, Siyi and Fouhey, David F},
  booktitle=CVPR,
  year={2022}
}

@article{kawana2021unsupervised,
  title={Unsupervised pose-aware part decomposition for 3d articulated objects},
  author={Kawana, Yuki and Mukuta, Yusuke and Harada, Tatsuya},
  journal={arXiv preprint arXiv:2110.04411},
  year={2021}
}

@inproceedings{sun2024opdmulti,
  title={Opdmulti: Openable part detection for multiple objects},
  author={Sun, Xiaohao and Jiang, Hanxiao and Savva, Manolis and Chang, Angel},
  booktitle=ThreeDV,
  year={2024},
}

@inproceedings{deng2024articulate,
  title={Articulate your NeRF: Unsupervised articulated object modeling via conditional view synthesis},
  author={Deng, Jianning and Subr, Kartic and Bilen, Hakan},
  booktitle=NIPS,
  year={2024}
}

@inproceedings{liu2023building,
  title={Building rearticulable models for arbitrary 3d objects from 4d point clouds},
  author={Liu, Shaowei and Gupta, Saurabh and Wang, Shenlong},
  booktitle=CVPR,
  year={2023}
}

@inproceedings{ma2023sim2real,
  title={Sim2Real 2: Actively Building Explicit Physics Model for Precise Articulated Object Manipulation},
  author={Ma, Liqian and Meng, Jiaojiao and Liu, Shuntao and Chen, Weihang and Xu, Jing and Chen, Rui},
  booktitle=ICRA,
  year={2023},
}

@inproceedings{guo2025articulatedgs,
  title={Articulatedgs: Self-supervised digital twin modeling of articulated objects using 3d gaussian splatting},
  author={Guo, Junfu and Xin, Yu and Liu, Gaoyi and Xu, Kai and Liu, Ligang and Hu, Ruizhen},
  booktitle=CVPR,
  year={2025}
}

@inproceedings{pekelny2008articulated,
  title={Articulated object reconstruction and markerless motion capture from depth video},
  author={Pekelny, Yuri and Gotsman, Craig},
  booktitle={CGF},
  year={2008},
}

@inproceedings{li2023imagemanip,
  title={Imagemanip: Image-based robotic manipulation with affordance-guided next view selection},
  author={Li, Xiaoqi and Wang, Yanzi and Shen, Yan and Iaroslav, Ponomarenko and Lu, Haoran and Wang, Qianxu and An, Boshi and Liu, Jiaming and Dong, Hao},
  booktitle={arXiv preprint arXiv:2310.09069},
  year={2023}
}

@inproceedings{wu2025afforddp,
  title={Afforddp: Generalizable diffusion policy with transferable affordance},
  author={Wu, Shijie and Zhu, Yihang and Huang, Yunao and Zhu, Kaizhen and Gu, Jiayuan and Yu, Jingyi and Shi, Ye and Wang, Jingya},
  booktitle=CVPR,
  year={2025}
}

@inproceedings{geng2023gapartnet,
  title={Gapartnet: Cross-category domain-generalizable object perception and manipulation via generalizable and actionable parts},
  author={Geng, Haoran and Xu, Helin and Zhao, Chengyang and Xu, Chao and Yi, Li and Huang, Siyuan and Wang, He},
  booktitle={Proceedings of the IEEE/CVF Conference on Computer Vision and Pattern Recognition},
  year={2023}
}

@article{huang2014occlusion,
  title={Occlusion-aware multi-view reconstruction of articulated objects for manipulation},
  author={Huang, Xiaoxia and Walker, Ian and Birchfield, Stan},
  journal={Robotics and Autonomous Systems},
  volume={62},
  number={4},
  pages={497--505},
  year={2014},
  publisher={Elsevier}
}
\end{document}